%% file: batch.tex
\documentclass[runningheads]{llncs}

\usepackage{microtype}
\usepackage{graphicx}
\usepackage{booktabs} 
\usepackage{vwcol}
\usepackage{hyperref}

\usepackage[linesnumbered,ruled,vlined]{algorithm2e}
\newcommand{\eat}[1]{}

\usepackage{caption}
\usepackage{subcaption}
\usepackage{amsmath}
\usepackage{paralist}
\input{math_commands.tex}

\begin{document}
%
\title{How do SGD hyperparameters in natural training affect adversarial robustness?}
\titlerunning{Role of SGD Hyperparameters in Adversarial Robustness}
%
\author{Sandesh Kamath\inst{1} \and
Amit Deshpande\inst{2} \and
K V Subrahmanyam\inst{1}}
%
%
\institute{Chennai Mathematical Institute, Chennai, India \and
Microsoft Research, Bengaluru, India\\
\email{{ksandeshk,kv}@cmi.ac.in}\\
\email{amitdesh@microsoft.com}}
%
\maketitle              
\begin{abstract}
Learning rate, batch size and momentum are three important hyperparameters in the SGD algorithm. It is known from the work of \cite{Jas17} that large batch size training of neural networks yields models which do not generalize well.  \cite{Yao18a} observe that large batch training yields models that have poor adversarial robustness. In the same paper, the authors train models with different batch sizes and compute the eigenvalues of the Hessian of loss function. They observe that as the batch size increases, the dominant eigenvalues of the Hessian become larger. They also show that both adversarial training and small-batch training leads to a drop in the dominant eigenvalues of the Hessian or lowering its spectrum. They combine adversarial training and second order information to come up with a new large-batch training algorithm and obtain robust models with good generalization. In this paper, we empirically observe the effect of the SGD hyperparameters on the accuracy and adversarial robustness of networks trained with unperturbed samples.  \cite{Jas17} considered training models with a fixed learning rate to batch size ratio. They observed that higher the ratio, better is the generalization. We observe that networks trained with constant learning rate to batch size ratio, as proposed in \cite{Jas17}, yield models which generalize well and also have almost constant adversarial robustness, independent of the batch size. We observe that momentum is more effective with varying batch sizes and a fixed learning rate than with constant learning rate to batch size ratio based SGD training.
\end{abstract}

\section{Introduction}
In machine learning it is important that models learned are robust to out-of-distribution data and to the data distribution the model is trained with. One way to achieve this robustness is data augmentation, and adversarial training is a special case. While there are studies relating the various hyperparameters e.g. \cite{Smith17},\cite{Jas17} for good accuracy, there is less attention towards tuning hyperparameters for simultaneous natural accuracy and adversarial robustness. In this work we initiate such a study by training models with a fixed learning rate to batch size ratio, as proposed in \cite{Jas17}. We evaluate the natural accuracy and adversarial robustness of models trained in this manner using only unperturbed inputs during the training phase.  We study the Hessian spectrum of models trained in this fashion as done in \cite{Yao18a} and report our observations.

Stochastic Gradient Descent (SGD) and its variants are the current workhorse for training neural network models. Hyperparameters like learning rate, batch size and  momentum play an important role in SGD converging to models which generalizes well. \cite{Smith17} and \cite{Hoffer17} study some relations between these hyperparameters and suggest rules for better training accuracy. In \cite{Goyal17} the authors give a rule that adjusts learning rate as a function of minibatch size. This results in a significant speedup in the time it takes to train networks with ImageNet in a distributed setting. 

It has been observed that there is a trade-off between generalization of a model and the batch size used in the training when the batch size is larger than 512. \cite{Keskar16} show that large batch training results in neural networks settling down at sharp minima and not being able to generalize well. They also propose a solution to handle the sharp minima issue along with better generalization.  On the other hand \cite{Dinh17} report that sharp minima on deeper networks can also generalize well. \cite{Jas17} show that maintaining a constant ratio between learning rate and batch size during training results in networks converging to a flatter minima leading to better generalization.


Neural networks achieve state-of-the-art results for many classification tasks. However, recent work of \cite{Szegedy13}, \cite{Biggio13} has exposed a serious vulnerability in neural network-based models that achieve state-of-the-art results on tasks such as object recognition. These models are known to be vulnerable to small, pixel-wise perturbations to the inputs. While these changes are almost imperceptible to the human eye neural networks grossly misclassify such perturbed data, even when they classify the unperturbed data correctly. \cite{Szegedy13} obtain these small perturbations using box-constrained L-BFGS, by maximizing the prediction error of the given model. In \cite{Goodfellow15} the authors propose a quicker method based on gradients, the Fast Gradient Sign Method (FGSM).  An FGSM adversarial perturbation of an input $x$ is given by $x' = x + \epsilon~ \text{sign}\left(\nabla_{x} J(\theta, x, y)\right)$. Here $y$ is the target, $\theta$ are the model parameters, and $J(\theta, x, y)$ is the loss function used to train the network. Subsequent work has introduced multi-step variants of FGSM, notably, an iterative method (BIM or $FGSM^k$) by \cite{Kurakin17} and Projected Gradient Descent (PGD) by \cite{Madry18}. On visual tasks, the adversarial perturbation must come from a set of images that are perceptually similar to a given image. In \cite{Goodfellow15} and \cite{Madry18} the authors study adversarial perturbations from an $\ell_{\infty}$-ball around the input $x$, namely, each pixel value is perturbed by a quantity within $[-\epsilon, + \epsilon]$.

A simple method to obtain a adversarially robust network would be to include the perturbed samples into the training process. This is referred to as \emph{adversarial training}. It is a special type of data augmentation technique where the data is augmented with adversarial perturbationi e.g. PGD while training. This method is expensive, but currently provides one of the most adversarially robust models. Some have tried a mixed approach. For example in \cite{Yao18a}, they initially train the network for 100 epochs to achieve good accuracy with unperturbed samples and then for 5-10 epochs do adversarial training with FGSM perturbated samples. While \cite{Yao18a} restrict their study to FGSM based training and FGSM based test accuracy, we have also included PGD based test results in Appendix for all the plots in our study, as PGD based attack is currently believed to be the strongest $\ell_{\infty}$-based adversarial attack.

A given network can be made robust either by explicit regularization like adversarial training, weight conditioning or by implicit regularization through hyperparameter tuning. Our work could be viewed as understanding the influence of hyperparameters like learning rate, batch size, momentum on the adversarial robustness of networks trained with natural samples.
Obtaining a naturally robust system without adversarial training is a desirable property. \cite{Sabour17} address this issue at the architecture level, without adversarial training. In \cite{Schmidt18} the authors study the sample complexity of adversarial generalization and claim that a much larger sample size would be needed to achieve adversarial robustness. \cite{Galloway18} observe that SGD with weight decay yields a robust network with better generalization than what can be achieved by adversarial training. \emph{Our focus is on the effect of hyperparameters of SGD on adversarial robustness, without weight decay.}

\cite{Yao18a} compute the Hessian spectrum of the parameter space of networks trained with large batch size and consistently observe that the top eigen value of the Hessian (the principle curvature) has a large magnitude. They empirically observe a direct correlation between the adversarial robustness of networks and principle curvature. They conclude that since large batch training leads to convergence to points with high principle curvature, the resulting models are not adversarially robust.

Our paper tries to understand the natural robustness of networks obtained by SGD hyperparameter tuning alone. Motivated by the works of \cite{Jas17} and \cite{Yao18a} we seek to understand the relation between the network weights obtained under various hyperparameter settings of SGD and the resulting networks FGSM/PGD adversarial robustness. For this study we use MNIST, Fashion MNIST and CIFAR-10 datasets. To compare with existing literature we use the M1 and C1 models from \cite{Yao18a}. The StdCNN model we use is described in Table \ref{stdcnn-table} and the ResNet18 model we use is from \cite{He16}.

\noindent \textbf{Our Results}

\noindent We make the following empirical observations.
\begin{itemize}
	\item [$\bullet$] Training models with a constant learning rate to batch size ratio not only leads to convergence to a flatter minima but also ensures that adversarial robustness does not degrade with increasing batch size during training.
	\item [$\bullet$] We show that the Hessian based analysis of \cite{Yao18a} does not always explain adversarial robustness in small vs large batch training. 
	\item [$\bullet$] We show that there are models which have higher Hessian spectrum when trained with large batch size yet have better adversarial accuracy when compared to models trained with small batch size.
	\item [$\bullet$] Adding momentum does help in converging to a flatter minima - this is empirically substantiated by a lowering  of the Hessian spectrum. Training with a larger momentum value leads to (in most cases) a more robust model than training with a smaller momentum value.
\end{itemize}


\subsection{SGD, its variants and hyperparameters}

The simplest gradient based training algorithm is batch gradient. Since all data does not fit into memory, running the vanilla batch gradient for each step becomes expensive because data needs to be brought into memory for gradient calculations. So, in practice, the algorithm used to train a neural network model to learn its free parameters $\theta$ by optimizing the loss function $L(\theta, x, y)$ is a version of gradient descent known as Stochastic Gradient Descent (SGD). SGD and its variants have been shown to converge to good local minima of $L$ (a non-convex function) that generalize well.
Another version of SGD used in practice is \textit{Mini-Batch Gradient Descent with Momentum} (Algorithm \ref{sgd:mbgdm}). In this formulation we can see an interplay between three hyperparameters of SGD, the learning rate($\eta$), batch size(b) and momentum($\gamma$). When $\gamma=0$ the formulation reduces to Mini-Batch Gradient Descent. When $\gamma=0$ and $b=1$ it becomes Stochastic Gradient Descent. When $\gamma=0$ and b=N we get the vanilla Batch Gradient Descent. To maintain a balance, mini-batch gradient method is used in practice. 

\begin{algorithm}[]
\SetAlgoLined
\SetKwInOut{Input}{Input}
\SetKwInOut{Output}{Output}
\Input{Loss fn $L(\theta, x, y)$, Momentum $\gamma$, Learning rate $\eta$, Batch size b, No. of batches $B$ = $\ceil{N/b}$.}
\Output{New Model Parameters $\theta$}
\SetKwFor{For}{for (}{) $\lbrace$}{$\rbrace$}
\tcp{Loop for number of batches}
\For{$i = 0;\ i < B;\ i = i + 1$}{
\tcp{Velocity vector used for Momentum calculation}
$\upsilon_{-1} = 0$\\
\tcp{Calculate the gradient for a batch}
\For{$j = 0;\ j < b;\ j = j + 1$}{
\tcp{Gradient Calculation and Updation}
$\upsilon_{i} =  \gamma.\upsilon_{i-1} - \nabla_\theta L (\theta, x^{(i+j)}, y^{(i+j)})$\\
$\theta = \theta + \eta . \upsilon_{i} $\\
}
}
 \caption{Mini-Batch Gradient Descent with Momentum}
\label{sgd:mbgdm}
\end{algorithm}

From the variants of SGD, we see that learning rate($\eta$) is an important hyperparameter. It has been observed that tuning the learning rate itself can aid in better convergence of the SGD algorithm to a minima that generalizes well. The common rule of thumb in tuning learning rate is to decay/decrease it as we train the network. In \cite{Smith17} the authors note that instead of the usual practice of decreasing the learning rate over the epochs, even increasing the batch size leads to network with similar test accuracy. This has the advantage of faster training time with fewer parameter updates. They even run experiments with an inverse relation between the batch size and momentum and show that this too leads to better training with a small drop in test accuracy.

So batch size is an important hyperparameter in the training of neural networks. Larger the batch size, faster is the training of the network. But available computation resources restrict the maximum batch size for training a given network architecture. \cite{Bengio12} and \cite{Bottou12} describe some common rules and tricks used in practice, for optimizing the parameters of neural network models with SGD. They show that small batch sizes $b=32$ and below are generally preferred. \cite{Hoffer17} note that training with larger batch size is good for speeding up the training time but they have a severe problem with generalization. They suggest training for a longer time to improve the generalization of the network. 

More recently, \cite{Jas17} show that maintaining a constant learning rate to batch size ratio aids the SGD algorithm to converge to a flat minima, which generalizes well to the test points. \cite{Yao18a} characterise the adversarial (FGSM) vulnerability of the network obtained with large batch training with the top eigen value of the Hessian. They notice a correlation between the adversarial robustness (FGSM) of the network and the top eigen value of the Hessian. But the caveat is that they do the study under a certain hyperparameter setting. They train the network using a varying learning rate to batch size ratio (refer Section \ref{sec:settings} for details of their hyperparameter setting). Therefore, we believe that a study of the role of hyperparameters like learning rate, batch size and the corresponding adversarial robustness is important. We empirically analyse the applicability of the theory of \cite{Jas17} to the understanding of the adversarial robustness of the network. We do this by systematically training the network with SGD using various hyperparameter settings, and checking the adversarial robustness of the resultant network. For each of these hyperparameter settings, we also study the role that momentum plays.

\begin{remark}
\emph{An important point to be noted in this paper is that at no point do we use adversarially perturbed samples in the training process. All the networks are trained with samples \textbf{without} adversarial perturbation.}
\end{remark}

\section{Hyperparameter settings, model architectures and data sets used.}
To get an understanding of the natural robustness of models, we trained models under various hyperparameter settings. For each of these settings we plotted the accuracy of the trained model and the adversarial accuracy of the trained model on test inputs which are adversarially perturbed. We emphasize again that we do not augment the training data with adversarially perturbed train inputs.
We performed our experiments on the models used in  \cite{Yao18a} and also on StdCNN (refer to Table \ref{stdcnn-table}) and ResNet18. \cite{Yao18a} only use FGSM based adversarial perturbations in their experiments for both training and testing. We provide results for both FGSM and PGD based test perturbations. In the next section we describe the hyperparameter settings used in our experiments and the models and data sets used.

\subsection{Details of Datasets and Model Parameters} \label{sec:settings}

The following hyperparameter settings were used during training. For each of these settings we performed our experiments with momentum set to 0.0, 0.2, 0.5 and 0.9.  In all our plots we clearly mention which hyperparameter settings have been used to obtain that plot.  In Section~\ref{zeromomentum}, we discuss our observations with momentum set to zero and in Section~\ref{nonzeromomentum}, we discuss our observations with non-zero momentum.
\begin{enumerate}
\item [(1)] LR (see \cite{Smith17}).  Here learning rate is fixed to 0.01 and batch size is varied and training is done with this setting for 100 epochs. Results for this hyperparameter setting are shown in light blue colour in all our plots.
\item [(2)] LR/BS(see \cite{Jas17}). Here learning rate to batch size ratio is kept constant. We set the ratio to 0.00015625 and training is done with this fixed setting for 100 epochs and varying batch sizes. Results for this hyperparameter setting are shown in purple in all our plots.
\item [(3)] For comparison with \cite{Yao18a}, we also train models using exactly the settings from their paper. Here the learning rate is set to 0.01 and momentum to 0.9, and learning rate is decayed by half after every 5 epochs, for a total of 100 epochs. Results for this hyperparameter setting are shown in red in all our plots, and we refer to this as Benchmark.

\end{enumerate}

Note that in the settings (1) and (2) above we do not use weight decay nor decay of learning rate. We fix the learning rate, batch size and momentum at the beginning of training with SGD, and no adaptive tuning is done to these settings during the training. 

For each of the hyperparameter settings and momenta values in the experiments above, we computed the largest eigen value of the Hessian with respect to model parameters, for varying batch size and plot those graphs.

\textbf{Data sets} MNIST dataset consists of $70,000$ images of $28 \times 28$ size, divided into $10$ classes. $55,000$ used for training, $5,000$ for validation and $10,000$ for testing. Fashion MNIST dataset consists of $70,000$ images of $28 \times 28$ size, divided into $10$ classes. $55,000$ used for training, $5,000$ for validation and $10,000$ for testing. CIFAR-10 dataset consists of $60,000$ images of $32 \times 32$ size, divided into $10$ classes. $40,000$ used for training, $10,000$ for validation and $10,000$ for testing. 

\textbf{Model Architectures} For the MNIST and Fashion MNIST based experiments we use the architectures M1 and StdCNN as given in the Table \ref{stdcnn-table}. 

For the CIFAR-10 based experiments we use the models C1 as given in Table \ref{stdcnn-table} and ResNet18 architecture as given in \cite{He16}. Input training data was augmented with random cropping and random horizontal flips by default.

Architectures M1 used for MNIST and Fashion MNIST experiments and C1 for CIFAR-10 experiments are as given in \cite{Yao18a}, which form the benchmark for comparison. 

All the PGD based attack results in the Appendix for the corresponding FGSM attack based plots in the paper were plotted with step size k = 40.

\begin{table}[h] 
\caption{Architectures used for experiments} \label{stdcnn-table}
\begin{center}
\resizebox{\linewidth}{!} {
\begin{tabular}{| l | l |}
\multicolumn{1}{c}{\bf Name}  &\multicolumn{1}{c}{\bf Structure}
\\ \hline 
StdCNN         & Conv(3,3,10) - Conv(3,3,10) - MP(2,2) - \\
               & Conv(3,3,20) - Conv(3,3,20) - MP(2,2) - \\
               & FC(50) - Dropout(0.5) - FC(10) - SM(10) \\ 
\hline
M1             & Conv(5,5,20) - Conv(5,5,20) - FC(500) - SM(10) \\ 
\hline
C1             & Conv(5,5,64) - MP(3,3) - BN - Conv(5,5,64) - \\
	       & MP(3,3) - BN - FC(384) - FC(192) - SM(10) \\ 
\hline
\end{tabular}
}
\end{center}
\end{table}

\section{Adversarial robustness using hyperparameter tuning in natural training}
\label{zeromomentum} In this section we report the results of our experiments when momentum is set to zero. For the benchmark alone we continue to use momentum 0.9.

We first verify that we get the same values and trends reported in \cite{Yao18a}. Our observations confirm the findings of \cite{Yao18a}. From the benchmark plots in Figures \ref{combined-mnist-lenet-fgsm-dist}(left) and \ref{combined-cifar10-c1-fgsm-dist}(left) it is clear that as the batch size increases the test accuracy decreases, and similarly the associated FGSM test accuracy also drops with increase in batch size. We were able to replicate the benchmark experiments on test accuracy. 
For  adversarial robustness using FGSM we observe the same trend reported in \cite{Yao18a}- that the accuracy drops with larger batch size. However, our accuracy values are different. 
It's clear from the top eigen value plots that as batch size increases principle curvature increases. This can be seen in the benchmark plots in Figure \ref{combined-mnist-lenet-fgsm-dist}(right), Figure \ref{combined-cifar10-c1-fgsm-dist}(right).

In Figure \ref{combined-mnist-lenet-fgsm-dist}(left) we plot the test accuracy as a function of batch size, and the adversarial accuracy as a function of batch size using FGSM attack with $\epsilon=0.3$ for model M1 on the MNIST dataset.  We plot this LR set to a constant (light blue line)  and LR/BS set to a constant (purple line).  For each setting of the hyperparameters, we computed the topmost eigen value as a function of batch size. This is plotted in Figure \ref{combined-mnist-lenet-fgsm-dist}(right). 

We repeated the above experiments on StdCNN and report the results in Figure \ref{combined-mnist-lenet-stdcnn-hessian}(left) and  Figure \ref{combined-mnist-lenet-stdcnn-hessian}(right). 

Figures \ref{combined-cifar10-c1-fgsm-dist} and \ref{combined-cifar10-stdcnn-hessian} show similar plots for CIFAR-10 on model C1 and ResNet18, respectively. We use $\epsilon=0.02$ for FGSM attack on CIFAR-10. The $\epsilon$ values are exactly as used by \cite{Yao18a} for their study. 

We also consider PGD attack on these models trained to classify MNIST. Figure \ref{combined-mnist-lenet-pgd-dist} (in Appendix \ref{app:sec:comp-yao}) shows the plots of accuracy and adversarial accuracy versus batch size for PGD attack on models M1 and StdCNN respectively, using with $\epsilon=0.3$ under various settings of hyperparameters. Figure \ref{combined-cifar10-c1-pgd-dist} (in Appendix \ref{app:sec:comp-yao}) shows the PGD attack plots for models C1 and ResNet18 classifying CIFAR-10, using $\epsilon=0.02$.

\begin{figure}[]
\begin{center}
\includegraphics[width=0.49\linewidth]{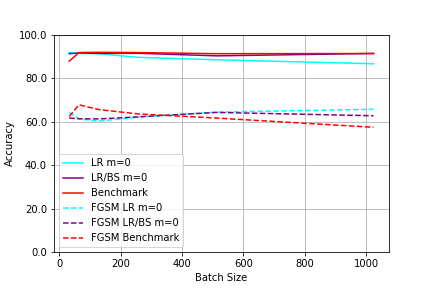}
\includegraphics[width=0.49\linewidth]{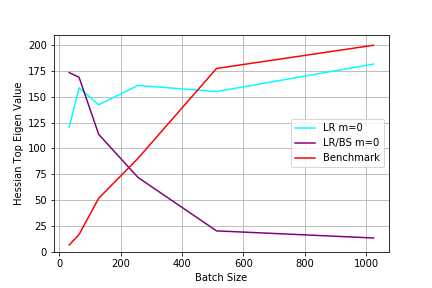}
\end{center}
\caption{On Fashion MNIST, (left) Test Accuracy of model M1 and using PGD attack with $\epsilon=0.3$. (right) $\lambda_{1}^{\theta}$ : Top Eigen value of Hessian of model M1. For LR, learning rate = 0.01. For LR/BS, ratio = 0.00015625. m=momentum.} 
\label{combined-fmnist-lenet-fgsm-dist}
\end{figure}

For models M1 and StdCNN we also performed the above experiments  when they were trained to classify Fashion MNIST. Figures \ref{combined-fmnist-lenet-fgsm-dist}(left), \ref{combined-fmnist-stdcnn-fgsm-dist}(left) plot the generalization as a function of batch size, and FGSM accuracy as a function of batch size for Fashion MNIST using $\epsilon=0.3$. In Figures \ref{combined-fmnist-lenet-fgsm-dist}(right),\ref{combined-fmnist-stdcnn-fgsm-dist}(right) we plot the top eigen value of the Hessian as a function of batch size.

\subsection{Observations}

\begin{figure}[]
\begin{center}
\includegraphics[width=0.49\linewidth]{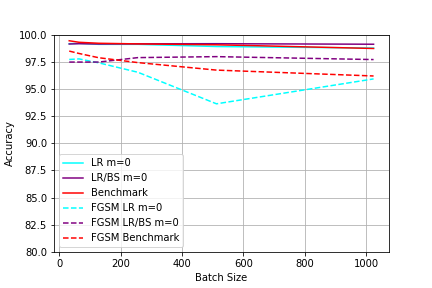}
\includegraphics[width=0.49\linewidth]{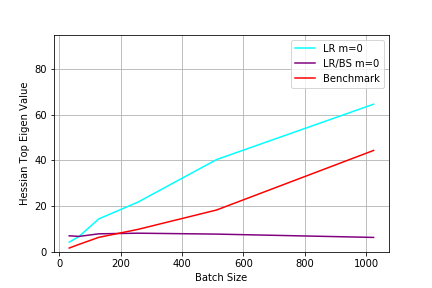}
\end{center}
\caption{On MNIST, (left) Test Accuracy of model M1 and using FGSM attack with $\epsilon=0.3$. (right) $\lambda_{1}^{\theta}$ : Top Eigen value of Hessian of model M1. For LR, learning rate = 0.01. For LR/BS, ratio = 0.00015625. m=momentum.} 
\label{combined-mnist-lenet-fgsm-dist}
\end{figure}

\begin{figure}[]
\begin{center}
\includegraphics[width=0.49\linewidth]{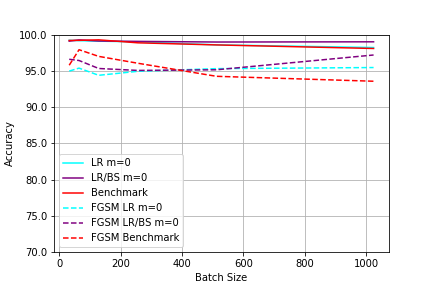}
\includegraphics[width=0.49\linewidth]{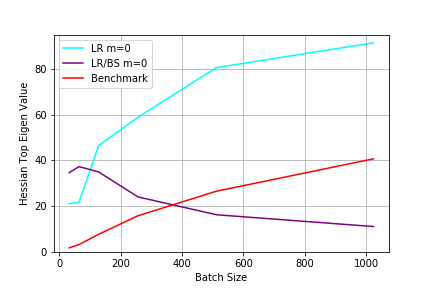}
\end{center}
\caption{On MNIST, (left) Test Accuracy of model StdCNN and using FGSM attack with $\epsilon=0.3$. (right) $\lambda_{1}^{\theta}$ : Top Eigen value of Hessian of model StdCNN. For LR, learning rate = 0.01. For LR/BS, ratio = 0.00015625. m=momentum.} 
\label{combined-mnist-lenet-stdcnn-hessian}
\end{figure}

\begin{figure}[]
\begin{center}
\includegraphics[width=0.49\linewidth]{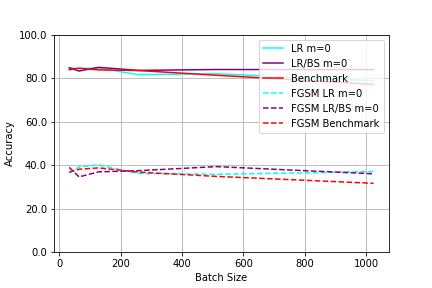}
\includegraphics[width=0.49\linewidth]{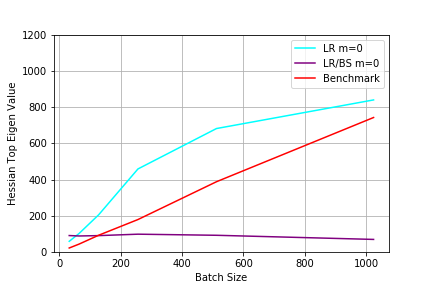}
\end{center}
\caption{On CIFAR-10, (left) Test Accuracy of model C1 and using FGSM attack with $\epsilon=0.02$. (right) $\lambda_{1}^{\theta}$ : Top Eigen value of Hessian of model C1. For LR, learning rate = 0.01. For LR/BS, ratio = 0.00015625. m=momentum.} 
\label{combined-cifar10-c1-fgsm-dist}
\end{figure}

\begin{figure}[]
\begin{center}
\includegraphics[width=0.49\linewidth]{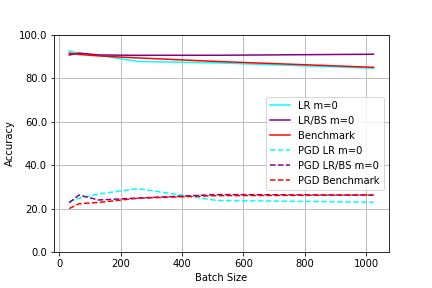}
\includegraphics[width=0.49\linewidth]{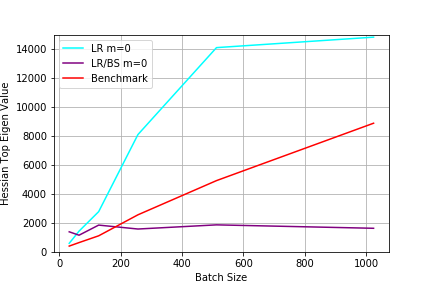}
\end{center}
\caption{On CIFAR-10, (left) Test Accuracy of model ResNet18 and using PGD attack with $\epsilon=0.02$. (right) $\lambda_{1}^{\theta}$ : Top Eigen value of Hessian of model ResNet18. For LR, learning rate = 0.01. For LR/BS, ratio = 0.00015625. m=momentum.} 
\label{combined-cifar10-stdcnn-hessian}
\end{figure}

\begin{itemize}
\item[(i)] Our observations confirm what \cite{Yao18a} observe - that as batch size increases natural accuracy drops.  We also confirm their observation that as batch size increases principle curvature increases. 
Our experiments also confirm the observations of \cite{Jas17}, that training with a constant LR/BS ratio leads to flatter minima. This can be observed from the purple lines in the top eigen value plots in Figure \ref{combined-mnist-lenet-fgsm-dist}(right), Figure\ref{combined-mnist-lenet-stdcnn-hessian}(right), Figure \ref{combined-cifar10-c1-fgsm-dist}(right) and Figure \ref{combined-cifar10-stdcnn-hessian}(right), which are almost flat. 

\item[(ii)] One major point to be noted here is that \emph{the purple line whether its generalization, FGSM/PGD accuracy or principle curvature of parameter space (top Hessian eigenvalue) there is very little variation across all models!} So with a constant LR/BS ratio training, neither the natural accuracy nor the adversarial accuracy suffer with an increase in batch-size during training.
This is not true for FGSM/PGD accuracy of models trained with constant LR nor is it true for benchmark, where one sees a drop in accuracy with increase in the batch-size during training.

The above observations continue to hold for  PGD attacks.  This can be seen from the purple line plots in Figure \ref{combined-mnist-lenet-pgd-dist} and Figure \ref{combined-cifar10-c1-pgd-dist} (both plots in Appendix \ref{app:sec:comp-yao}).

\item[(iii)] \cite{Yao18a} conclude that increased curvature results in a decrease in adversarial robustness - we show that it does not necessarily hold for all models and all datasets. We give details about this in Section~\ref{sec:counter}.

\end{itemize}
\eat{\begin{remark}
Our comparision with \cite{Yao18a} is only for FGSM robustness.  However, we have also given PGD attack based robustness plots..
\end{remark}
}

\clearpage
\subsection{Counter Example to the Hessian based analysis of Yao et al.\cite{Yao18a}} \label{sec:counter}
\begin{figure}[!h]
\begin{center}
\includegraphics[width=0.49\linewidth]{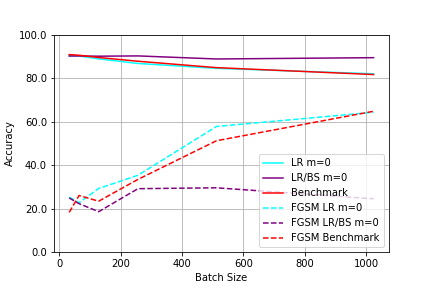}
\includegraphics[width=0.49\linewidth]{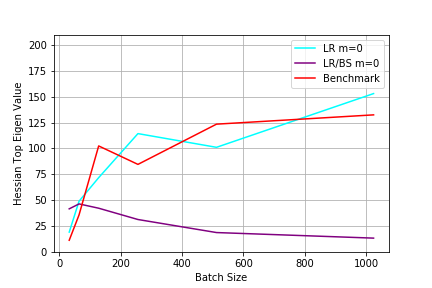}
\end{center}
\caption{On Fashion MNIST, (left) Test Accuracy of model StdCNN and using PGD attack with $\epsilon=0.3$. (right) $\lambda_{1}^{\theta}$ : Top Eigen value of Hessian of model StdCNN. For LR, learning rate = 0.01. For LR/BS, ratio = 0.00015625. m=momentum.} 
\label{combined-fmnist-stdcnn-fgsm-dist}
\end{figure}

We give an example of a model which when trained with large batch has higher Hessian spectrum and also higher adversarial robustness compared to small batch training. For this experiment we use the network given in Table\ref{stdcnn-table} trained on Fashion MNIST. We train the network with a fixed learning rate. We observe in Figure \ref{combined-fmnist-stdcnn-fgsm-dist}(left) that the FGSM accuracy increases with batch size (the light blue line in the plots). As expected the curvature also increases. This can be seen in Figure \ref{combined-fmnist-stdcnn-fgsm-dist}(right). 
On this model we see the same behaviour with respect to the PGD attack. This can be seen from Figure \ref{combined-fmnist-stdcnn-pgd-dist}(right) (in Appendix \ref{app:sec:counter}) where the PGD accuracy also increases with increasing batch size. \emph{So this clearly indicates that an increase in Hessian spectrum alone cannot explain the change in adversarial robustness of neural networks.} 

\section{The effect of momentum on adversarial robustness} \label{nonzeromomentum}

We now analyse the role of momentum in the two hyperparameter settings, LR and LR/BS that we used in Section~\ref{zeromomentum} to compare with Benchmark. For each of these settings, we train each model with momentum set to 0.0, 0.2, 0.5 and 0.9, and plot the natural accuracy on the test data as a function of batch size and the adversarial accuracy as a function of batch size when the test inputs are adversarially perturbed using FGSM and PGD (in Appendix \ref{app:sec:m}). We also plot the top eigen values of the Hessian as a function of batch size in each of these settings.

Its clear from all the Hessian eigenvalue plots in Figures \ref{combined-mnist-lenet-momentum-fgsm-dist}, \ref{combined-mnist-lenet-stdcnn-hessian-momentum}, \ref{combined-lenet-momentum-fgsm-dist-lrbs}, \ref{combined-mnist-lenet-stdcnn-hessian-lrbs} for MNIST and Figures \ref{combined-c1-momentum-fgsm-dist}, \ref{combined-cifar10-stdcnn-hessian-momentum}, \ref{combined-cifar10-c1-momentum-fgsm-dist-lrbs}, \ref{combined-cifar10-stdcnn-hessian-lrbs} for CIFAR-10 that \emph{in both the set ups LR and LR/BS, accuracy improves with increased momentum. Furthermore, increasing momentum  leads to  convergence to points with a lower Hessian spectrum.}
 
\subsection{Effect of Momentum with LR} \label{subsec:lr}
\begin{figure}[]
\begin{center}
\includegraphics[width=0.49\linewidth]{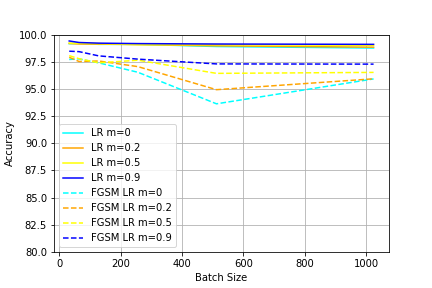}
\includegraphics[width=0.49\linewidth]{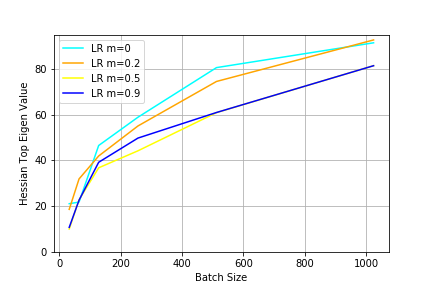}
\end{center}
\caption{On MNIST, (left) Test Accuracy of model M1 and using FGSM attack with $\epsilon=0.3$. (right) $\lambda_{1}^{\theta}$ : Top Eigen value of Hessian of model M1. For LR, learning rate = 0.01. m=momentum.} 
\label{combined-mnist-lenet-momentum-fgsm-dist}
\end{figure}

\begin{figure}[]
\begin{center}
\includegraphics[width=0.49\linewidth]{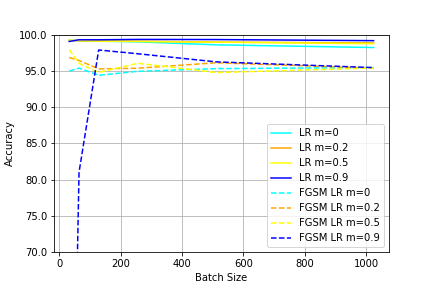}
\includegraphics[width=0.49\linewidth]{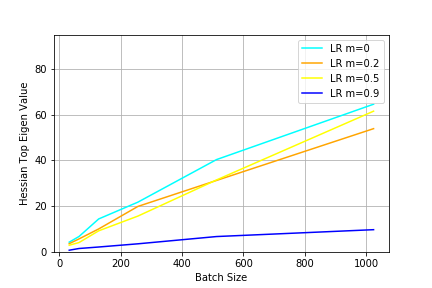}
\end{center}
\caption{On MNIST, (left) Test Accuracy of model StdCNN and using PGD attack with $\epsilon=0.3$. (right) $\lambda_{1}^{\theta}$ : Top Eigen value of Hessian of model StdCNN. For LR, learning rate = 0.01. m=momentum.} 
\label{combined-mnist-lenet-stdcnn-hessian-momentum}
\end{figure}

\begin{figure}[]
\begin{center}
\includegraphics[width=0.49\linewidth]{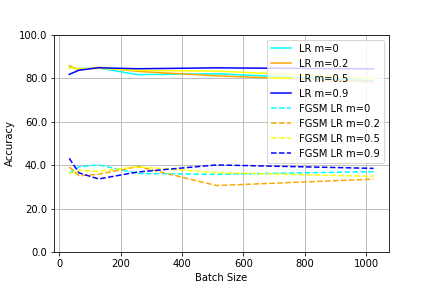}
\includegraphics[width=0.49\linewidth]{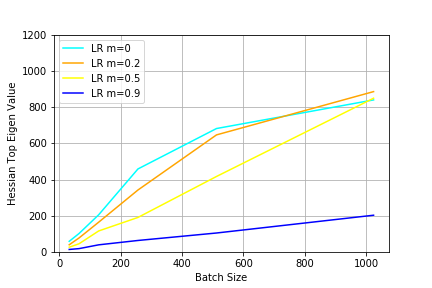}
\end{center}
\caption{On CIFAR-10, (left) Test Accuracy of model C1 and using PGD attack with $\epsilon=0.02$. (right) $\lambda_{1}^{\theta}$ : Top Eigen value of Hessian of model C1. For LR, learning rate = 0.01. m=momentum.} 
\label{combined-c1-momentum-fgsm-dist}
\end{figure}

\begin{figure}[]
\begin{center}
\includegraphics[width=0.49\linewidth]{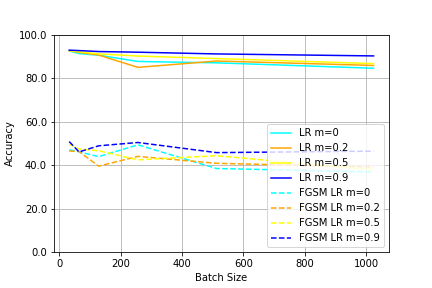}
\includegraphics[width=0.49\linewidth]{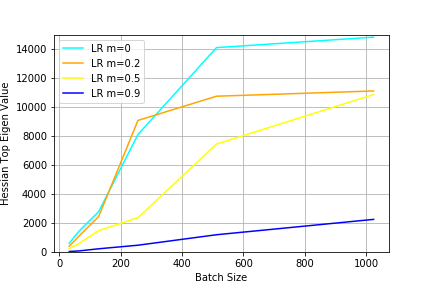}
\end{center}
\caption{On CIFAR-10, (left) Test Accuracy of model ResNet18 and using FGSM attack with $\epsilon=0.02$. (right) $\lambda_{1}^{\theta}$ : Top Eigen value of Hessian of model ResNet18. For LR, learning rate = 0.01. m=momentum.} 
\label{combined-cifar10-stdcnn-hessian-momentum}
\end{figure}

For a finer analysis of the impact of momentum with fixed LR, we plot the natural accuracy and adversarial accuracy with momentum values set to 0.0, 0.2, 0.5 and 0.9 for the models M1 and StdCNN on MNIST and models C1 and ResNet18 on CIFAR-10. Figures \ref{combined-mnist-lenet-momentum-fgsm-dist}(left), \ref{combined-mnist-lenet-stdcnn-hessian-momentum}(left), \ref{combined-c1-momentum-fgsm-dist}(left) and \ref{combined-cifar10-stdcnn-hessian-momentum}(left) show the generalization trend and, as expected, with larger momentum there is better generalization. This in turn leads to better FGSM adversarial robustness as seen in Figures \ref{combined-mnist-lenet-momentum-fgsm-dist}(left) and \ref{combined-mnist-lenet-stdcnn-hessian-momentum}(left) for MNIST and Figures \ref{combined-c1-momentum-fgsm-dist}(left) and \ref{combined-cifar10-stdcnn-hessian-momentum}(left) for CIFAR-10. For PGD robustness plots refer to Figure \ref{combined-mnist-lenet-momentum-pgd-dist} for MNIST and Figure \ref{combined-c1-momentum-pgd-dist} for CIFAR-10. Similarly in Figures \ref{combined-mnist-lenet-momentum-fgsm-dist}(right), \ref{combined-mnist-lenet-stdcnn-hessian-momentum}(right), \ref{combined-c1-momentum-fgsm-dist}(right) and \ref{combined-cifar10-stdcnn-hessian-momentum}(right) its seen that the curvature reduces with larger momentum. The PGD based plots are in Appendix \ref{app:subsec:lr}. 

\clearpage

\subsection{Effect of Momentum with LR/BS} \label{subsec:lrbs}
\begin{figure}[]
\begin{center}
\includegraphics[width=0.49\linewidth]{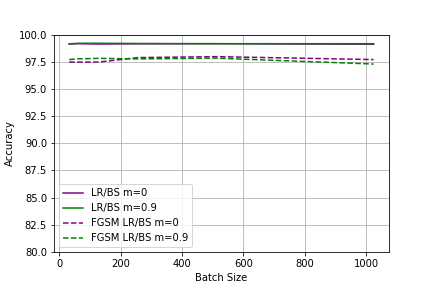}
\includegraphics[width=0.49\linewidth]{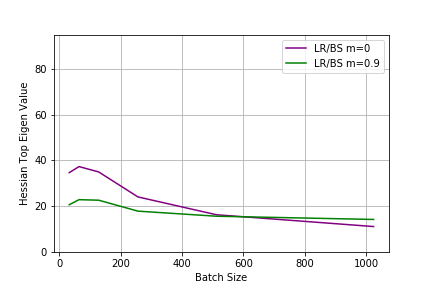}
\end{center}
\caption{On MNIST, (left) Test Accuracy of model M1 and using FGSM attack with $\epsilon=0.3$. (right) $\lambda_{1}^{\theta}$ : Top Eigen value of Hessian of model M1. For LR/BS, ratio = 0.00015625. m=momentum.} 
\label{combined-lenet-momentum-fgsm-dist-lrbs}
\end{figure}

\begin{figure}[]
\begin{center}
\includegraphics[width=0.49\linewidth]{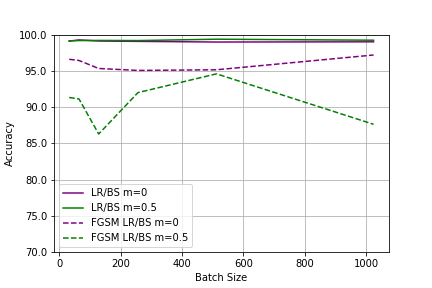}
\includegraphics[width=0.49\linewidth]{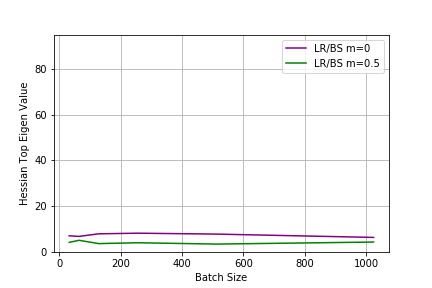}
\end{center}
\caption{On MNIST, (left) Test Accuracy of model StdCNN and using FGSM attack with $\epsilon=0.3$. (right) $\lambda_{1}^{\theta}$ : Top Eigen value of Hessian of model StdCNN. For LR/BS, ratio = 0.00015625. m=momentum.} 
\label{combined-mnist-lenet-stdcnn-hessian-lrbs}
\end{figure}

\begin{figure}[]
\begin{center}
\includegraphics[width=0.49\linewidth]{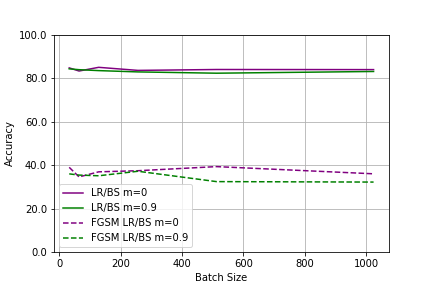}
\includegraphics[width=0.49\linewidth]{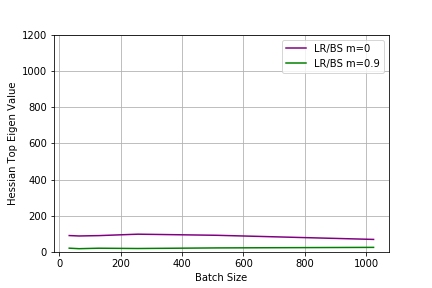}
\end{center}
\caption{On CIFAR-10, (left) Test Accuracy of model C1 and using FGSM attack with $\epsilon=0.02$. (right) $\lambda_{1}^{\theta}$ : Top Eigen value of Hessian of model C1. For LR/BS, ratio = 0.00015625. m=momentum.} 
\label{combined-cifar10-c1-momentum-fgsm-dist-lrbs}
\end{figure}

\begin{figure}[]
\begin{center}
\includegraphics[width=0.49\linewidth]{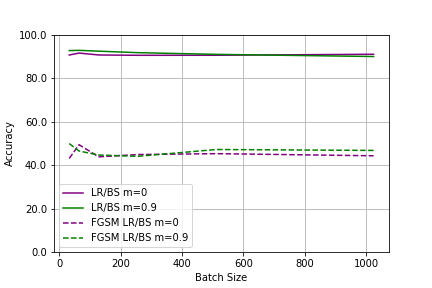}
\includegraphics[width=0.49\linewidth]{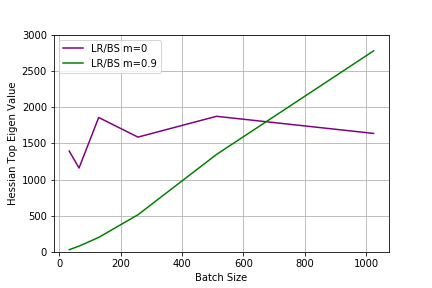}
\end{center}
\caption{On CIFAR-10, (left) Test Accuracy of model ResNet18 and using FGSM attack with $\epsilon=0.02$. (right) $\lambda_{1}^{\theta}$ : Top Eigen value of Hessian of model ResNet18. For LR/BS, ratio = 0.00015625. m=momentum.} 
\label{combined-cifar10-stdcnn-hessian-lrbs}
\end{figure}

For a finer analysis of the impact of momentum with fixed LR/BS  we plot the natural accuracy and adversarial accuracy with momentum values set to 0 and 0.5 or 0.9 for the models M1 and StdCNN on MNIST and models C1 and ResNet18 on CIFAR-10. In Figures \ref{combined-lenet-momentum-fgsm-dist-lrbs}, \ref{combined-mnist-lenet-stdcnn-hessian-lrbs}, \ref{combined-cifar10-c1-momentum-fgsm-dist-lrbs} and \ref{combined-cifar10-stdcnn-hessian-lrbs} its seen that in most cases the curvature reduces with larger momentum, but as compared to fixed LR training there is no significant role of momentum with constant learning rate and batch size ratio.  As the mild change in curvature or generalization does not always convert into better generalization or adversarial robustness as seen in Figures \ref{combined-lenet-momentum-fgsm-dist-lrbs}(left), \ref{combined-mnist-lenet-stdcnn-hessian-lrbs}(left), \ref{combined-cifar10-c1-momentum-fgsm-dist-lrbs}(left) and \ref{combined-cifar10-stdcnn-hessian-lrbs}(left) for FGSM attack or Figures \ref{combined-lenet-momentum-pgd-dist-lrbs} and \ref{combined-cifar10-c1-momentum-pgd-dist-lrbs} for PGD attack. The PGD based plots are in Appendix \ref{app:subsec:lrbs}. In this section we use fewer values of the momentum for the study as from the plots it is evident that momentum in the LR/BS setting doesnot play a significant role in the training and the resultant accuracy and adversarial robustness of the network.


\clearpage
\section{Conclusion}
We show how the modelling of SGD by \cite{Jas17} and the Hessian spectrum can help understand the weight space and its adversarial properties. We also see how momentum plays a role in reducing the spectrum of the parameters irrespective of the ratio maintained between learning rate and batch size. We believe the paper in its current form tries to understand the role of hyperparameters and the resultant networks robustness without any perturbed input. This would be necessary to gauge the impact of adversarial training on top of it and could aid in adapting the hyperparameters for adversarial training. It also opens the possibility of networks achieving adversarial robustness without adversarial training and only hyperparameter tuning.

\bibliographystyle{splncs04}
\bibliography{batch}

\appendix
\noindent \textbf{\large Appendix} \label{appendix:pgd}
\\
In this appendix we present the PGD based results for all the FGSM based plots in the paper. We present the plots corresponding to the respective sections in the paper in order.

\section{Adversarial robustness using hyperparameter tuning in natural training} \label{app:sec:comp-yao}
\begin{figure}[]
\begin{center}
\includegraphics[width=0.49\linewidth]{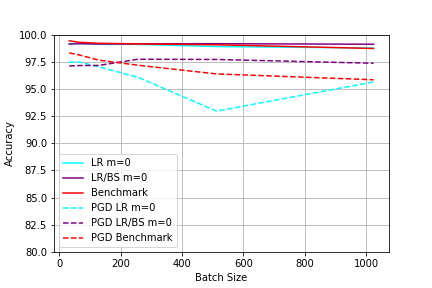}
\includegraphics[width=0.49\linewidth]{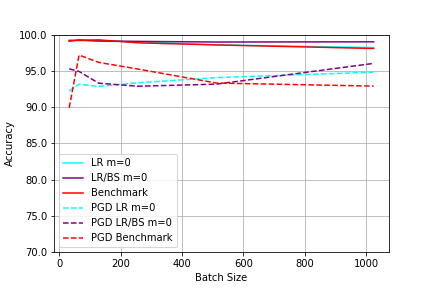}
\end{center}
\caption{Test Accuracy of models (left) M1 (right) StdCNN trained with MNIST and using PGD attack with $\epsilon=0.3$. For LR, learning rate = 0.01. For LR/BS, ratio = 0.00015625. m=momentum.} 
\label{combined-mnist-lenet-pgd-dist}
\end{figure}

\begin{figure}[]
\begin{center}
\includegraphics[width=0.49\linewidth]{plots/combined_lenet_fmnist_batch_gen_fgsm_notitle.png}
\includegraphics[width=0.49\linewidth]{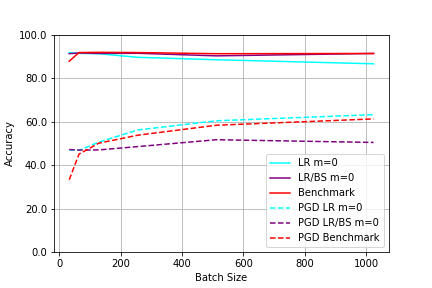}
\end{center}
\caption{Test Accuracy of M1 trained with Fashion MNIST and using (left) FGSM (right) PGD attack with $\epsilon=0.3$. For LR, learning rate = 0.01. For LR/BS, ratio = 0.00015625. m=momentum.} 
\label{combined-fmnist-lenet-pgd-dist}
\end{figure}

\begin{figure}[]
\begin{center}
\includegraphics[width=0.49\linewidth]{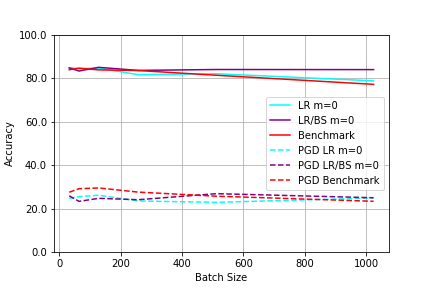}
\includegraphics[width=0.49\linewidth]{plots/combined_stdcnn_cifar10_batch_gen_pgd_notitle.png}
\end{center}
\caption{Test Accuracy of models (left) C1 (right) ResNet18 trained with CIFAR-10 and using PGD attack with $\epsilon=0.02$. For LR,learning rate = 0.01. For LR/BS, ratio = 0.00015625. m=momentum.} 
\label{combined-cifar10-c1-pgd-dist}
\end{figure}
\clearpage

\subsection{Counter Example to the Hessian based analysis of Yao et al.\cite{Yao18a}} \label{app:sec:counter}
\begin{figure}[!h]
\begin{center}
\includegraphics[width=0.49\linewidth]{plots/combined_stdcnn_fmnist_batch_gen_fgsm_notitle.png}
\includegraphics[width=0.49\linewidth]{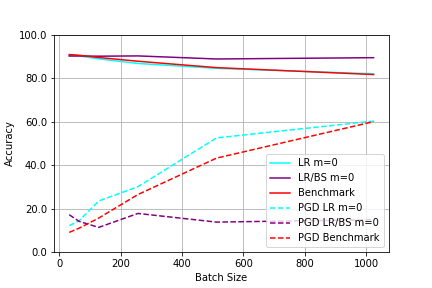}
\end{center}
\caption{Test Accuracy of StdCNN trained with Fashion MNIST and using (left) FGSM (right) PGD attack with $\epsilon=0.3$. For LR, learning rate = 0.01. For LR/BS, ratio = 0.00015625. m=momentum.} 
\label{combined-fmnist-stdcnn-pgd-dist}
\end{figure}

\section{The effect of momentum on adversarial robustness} \label{app:sec:m}
\subsection{Effect of Momentum with LR} \label{app:subsec:lr}
\begin{figure}[!h]
\begin{center}
\includegraphics[width=0.49\linewidth]{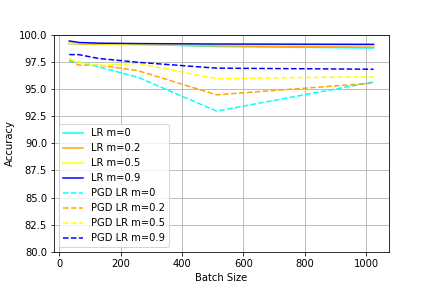}
\includegraphics[width=0.49\linewidth]{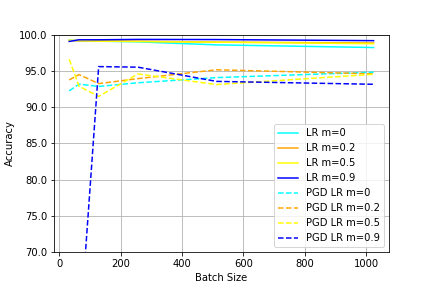}
\end{center}
\caption{Test Accuracy of models (left) M1 (right) StdCNN trained with MNIST and using PGD attack with $\epsilon=0.3$. For LR, learning rate = 0.01. m=momentum.} 
\label{combined-mnist-lenet-momentum-pgd-dist}
\end{figure}
\clearpage

\begin{figure}[!h]
\begin{center}
\includegraphics[width=0.49\linewidth]{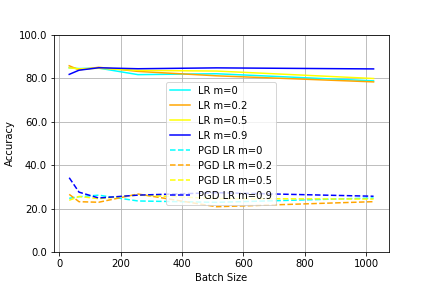}
\includegraphics[width=0.49\linewidth]{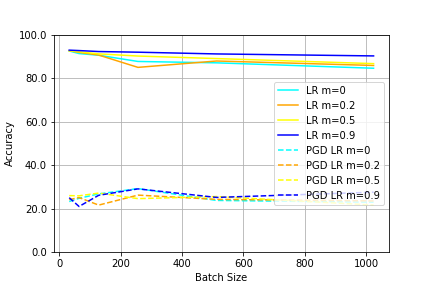}
\end{center}
\caption{Test Accuracy of models (left) C1 (right) ResNet18 trained with CIFAR-10 and using PGD attack with $\epsilon=0.02$. For LR,learning rate = 0.01. m=momentum.} 
\label{combined-c1-momentum-pgd-dist}
\end{figure}

\subsection{Effect of Momentum with LR/BS} \label{app:subsec:lrbs}
\begin{figure}[h!]
\begin{center}
\includegraphics[width=0.49\linewidth]{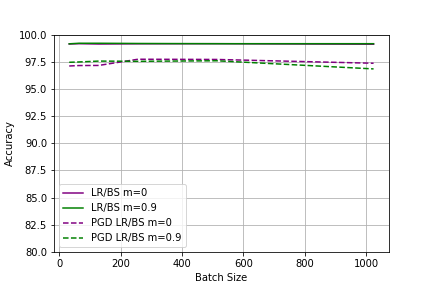}
\includegraphics[width=0.49\linewidth]{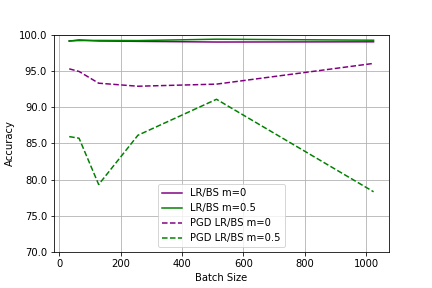}
\end{center}
\caption{Test Accuracy of models (left) M1, (right) StdCNN trained with MNIST and using PGD attack with $\epsilon=0.3$. For LR/BS, ratio = 0.00015625.} 
\label{combined-lenet-momentum-pgd-dist-lrbs}
\end{figure}

\begin{figure}[h!]
\begin{center}
\includegraphics[width=0.49\linewidth]{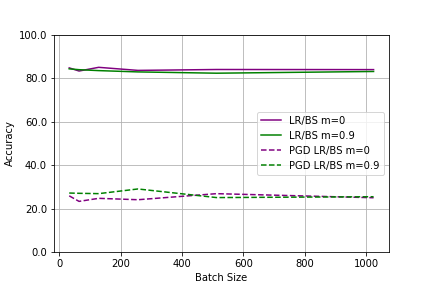}
\includegraphics[width=0.49\linewidth]{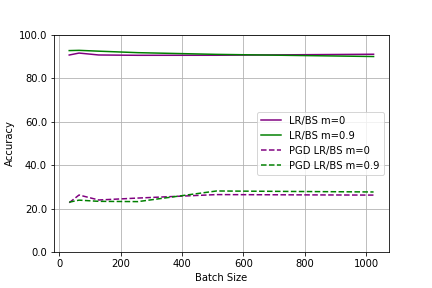}
\end{center}
\caption{Test Accuracy of models (left) C1 (right) ResNet18 trained with CIFAR-10 and using PGD attack with $\epsilon=0.02$. For LR/BS, ratio = 0.00015625.} 
\label{combined-cifar10-c1-momentum-pgd-dist-lrbs}
\end{figure}

\end{document}

%% file: math_commands.tex
\usepackage{amsmath,amsfonts,bm}

\def\eqref#1{equation~\ref{#1}}

\def\ceil#1{\lceil #1 \rceil}

\def\1{\bm{1}}

\newcommand{\test}{\mathcal{D_{\mathrm{test}}}}

\DeclareMathAlphabet{\mathsfit}{\encodingdefault}{\sfdefault}{m}{sl}
\SetMathAlphabet{\mathsfit}{bold}{\encodingdefault}{\sfdefault}{bx}{n}

